\documentclass{article}

\usepackage[preprint]{corl_2026} 
\usepackage{booktabs}
\usepackage{graphicx}
\usepackage{wrapfig}
\usepackage{float} 
\usepackage{pifont}
\usepackage{amsmath,amssymb}
\providecommand{\wrapfill}{}
\newcommand{\cmark}{\ding{51}}
\newcommand{\xmark}{\ding{55}}
\title{EgoAERO: Learning Dexterous Manipulation from a Single Egocentric Video without Object Assets}

\author{
\textbf{Yichen Niu\textsuperscript{1,2}, Haoran Lv\textsuperscript{1}, Xinrui Zhang\textsuperscript{1}, Xueyao Wan\textsuperscript{1}, Shiyu Gao\textsuperscript{1}}\\
\textbf{Ying Ai\textsuperscript{1}, Hui Xu\textsuperscript{1}, Yongqi Hu\textsuperscript{1}, Hengyi Zhang\textsuperscript{3}, Yang Xie\textsuperscript{3}}\\
\textbf{Zhaxizhuoma\textsuperscript{4,5}, Yue Zhao\textsuperscript{1}, Zhenshan Bing\textsuperscript{6}, Yan Ding\textsuperscript{2,7,8}, Jianxing Liu\textsuperscript{1,*}}\\[0.4em]
\normalfont\textsuperscript{1}School of Astronautics, Harbin Institute of Technology\\
\normalfont\textsuperscript{2}Lumos Robotic\\
\normalfont\textsuperscript{3}Suzhou Research Institute, Harbin Institute of Technology\\
\normalfont\textsuperscript{4}Shanghai Jiao Tong University \quad
\normalfont\textsuperscript{5}Shanghai AI Lab\\
\normalfont\textsuperscript{6}Nanjing University \quad
\normalfont\textsuperscript{7}Xi'an Jiaotong-Liverpool University \quad
\normalfont\textsuperscript{8}Fudan University\\
\normalfont\textsuperscript{*}Corresponding author
}

\begin{document}
\raggedbottom

\maketitle
\vspace{-2.5em}

\begin{figure}[H]
    \centering
    \includegraphics[width=0.9\linewidth]{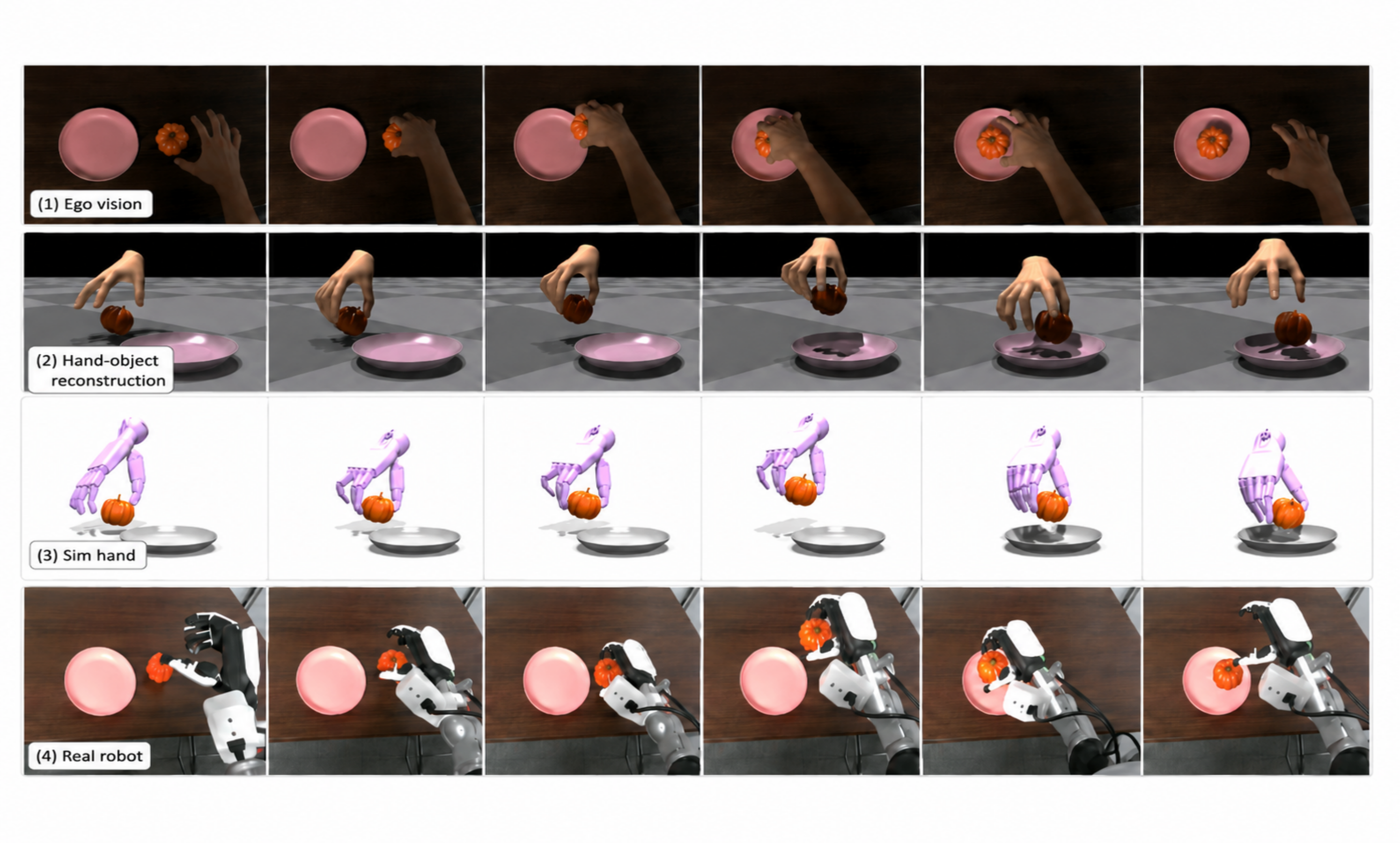}
    \vspace{-0.7em}
    \caption{End-to-end overview of EgoAERO. Starting from a single egocentric RGB-D human demonstration, EgoAERO reconstructs contact-consistent hand-object trajectories without object assets, transfers them to a simulated dexterous hand through two-stage policy learning, and executes the learned manipulation behavior on a real-world robot.}
    \label{fig:teaser}
\end{figure}

\begin{abstract}
    Egocentric RGB-D videos offer a natural source of human dexterous manipulation demonstrations, but existing data is difficult to use for robot learning because object pose, geometry, and contact information are often missing or require pre-scanned object assets. We present EgoAERO, the first framework that learns dexterous manipulation from a single egocentric RGB-D human demonstration without object assets. EgoAERO reconstructs contact-consistent hand-object trajectories through asset-free object tracking and reconstruction, ego motion compensation, and adaptive contact optimization, then converts them into robot policies using two-stage residual learning. We further introduce an online quality assessment mechanism and construct EgoDex-R, a large-scale egocentric dataset with 4.3M RGB-D frames for dexterous policy learning. Simulation and real-world experiments show that EgoAERO enables single-demonstration dexterous manipulation and achieves downstream performance close to CAD-based reconstructions on HOI4D.
\end{abstract}

\keywords{Dexterous manipulation, Egocentric RGB-D demonstration, Asset-free reconstruction} 


\section{Introduction}
	
Natural human demonstrations are becoming an important source for training dexterous manipulation policies. Compared with teleoperation, UMI-style interfaces~\citep{chi2024universal,zhaxizhuoma2025fastumi}, or data gloves, egocentric RGB-D devices can collect natural hand motions, object interactions, and contact-rich manipulation in everyday environments with minimal intrusion. This makes ego demonstrations a promising data source for scalable dexterous robot learning.

However, existing ego data is still hard to use directly for policy learning. First, large-scale egocentric datasets such as EPIC-KITCHENS~\citep{damen2018epic}, Ego4D~\citep{grauman2022ego4d}, and EgoDex~\citep{hoque2025egodex} provide rich videos or hand motion, but usually lack the manipulated object's 6-DoF pose, geometry, and contact state, which are needed for object-conditioned rewards, contact constraints, and replayable tasks. Second, datasets with 3D hand-object annotations, including H2O~\citep{kwon2021h2o}, HOI4D~\citep{liu2022hoi4d}, HOT3D~\citep{banerjee2025hot3d}, HO-3D~\citep{hampali2020hannotate}, DexYCB~\citep{chao2021dexycb}, OakInk~\citep{yang2022oakink}, and ARCTIC~\citep{fan2023arctic}, often rely on CAD models, scanned meshes, multi-view capture, or known object assets. This limits their scalability to arbitrary daily objects and natural collection settings.

We propose \textbf{EgoAERO, the first framework that converts a single egocentric RGB-D human demonstration into an executable dexterous manipulation policy without object assets} (Fig.~\ref{fig:teaser}). EgoAERO first reconstructs structured hand-object trajectories from raw ego RGB-D data by combining lightweight MLLM semantic initialization, asset-free object tracking and reconstruction under hand occlusion, ego motion compensation, and adaptive contact optimization. It then trains a two-stage residual policy: a hand-tracking policy learns to follow the reconstructed human hand motion, and a residual policy uses object pose and contact feedback to produce executable dexterous manipulation.

We also design an online ego data quality assessment mechanism and build \textbf{EgoDex-R}. Simulation and real-world experiments show that EgoAERO learns executable manipulation from natural ego demonstrations, and HOI4D comparisons show that its asset-free reconstructions achieve downstream policy learning performance close to CAD-based reconstructions. Our main contributions are:
\begin{list}{$\bullet$}{\setlength{\leftmargin}{1.2em}\setlength{\labelwidth}{0.6em}\setlength{\labelsep}{0.6em}\setlength{\itemsep}{0.1em}\setlength{\parsep}{0pt}\setlength{\topsep}{0.2em}}
    \item \textbf{EgoAERO}: the first framework for learning dexterous manipulation from a single ego RGB-D video without object assets.
    \item An asset-free ego hand-object reconstruction pipeline with robust object tracking, geometry reconstruction, ego motion compensation, and adaptive contact optimization.
    \item An online ego quality assessment mechanism and \textbf{EgoDex-R}, a large-scale dataset with 4.3M RGB-D frames for dexterous policy learning.
\end{list}
\section{EgoAERO Method}
\begin{figure}
    \centering
    \includegraphics[width=1\linewidth]{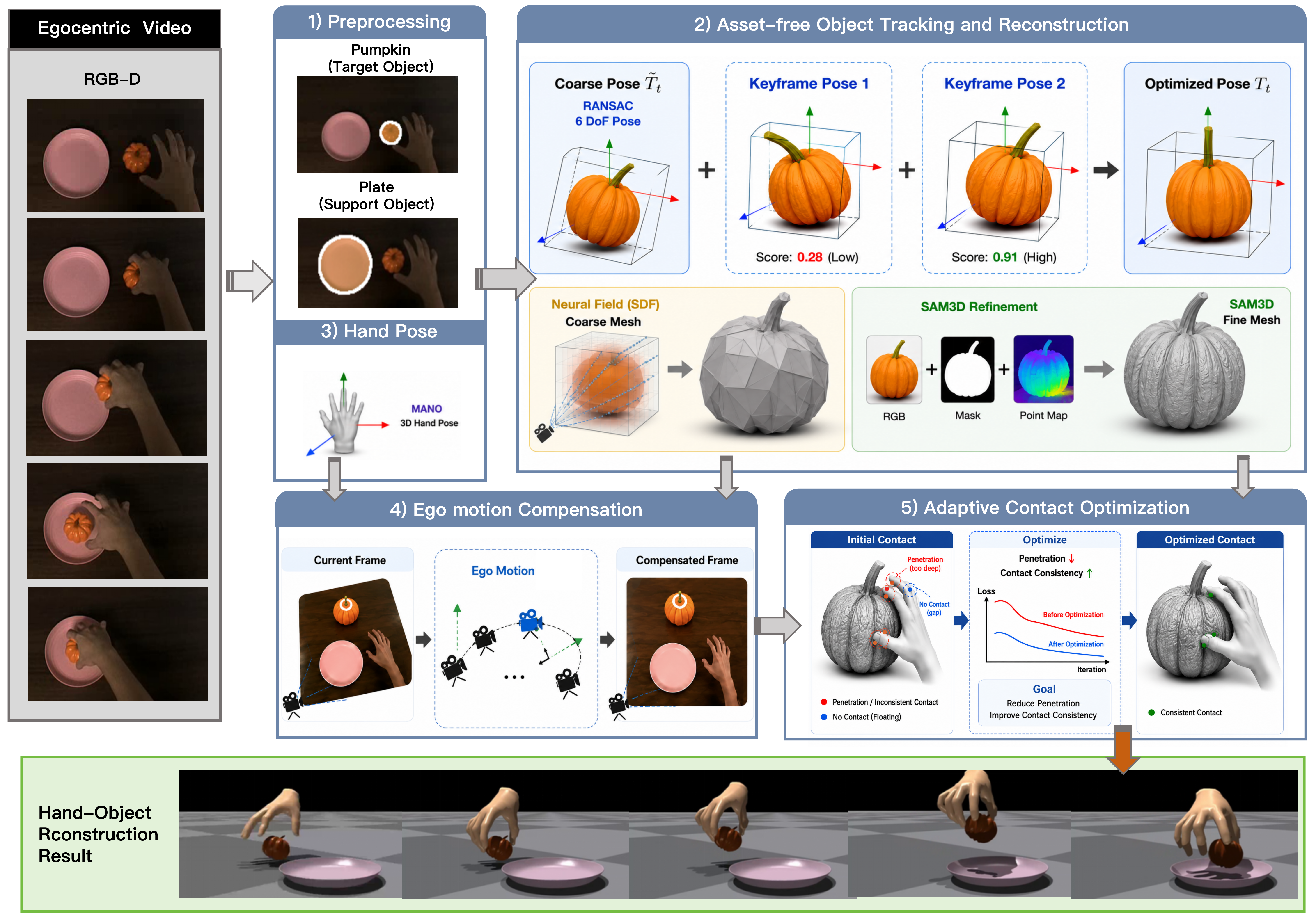}
    \vspace{-0.7em}
    \caption{Overview of asset-free egocentric hand-object reconstruction. Given a single ego RGB-D video, EgoAERO first performs MLLM-guided semantic preprocessing to identify the manipulated object and obtain segmentation prompts. It then reconstructs the unknown object without CAD assets through coarse pose initialization, keyframe memory-pool pose optimization, and neural-field-guided coarse-to-fine mesh reconstruction. The reconstructed object state is combined with RGB-D corrected hand pose, ego motion compensation, and adaptive contact optimization to produce contact-consistent hand-object trajectories for downstream robot learning.}
    \label{fig:aero_reconstruction}
\end{figure}

\subsection{Asset-free Egocentric Hand-Object Reconstruction}
\label{sec:EgoAERO}
Fig.~\ref{fig:aero_reconstruction} illustrates the overall pipeline of EgoAERO's asset-free egocentric hand-object reconstruction module.
\subsubsection{MLLM-Guided Semantic Preprocessing for Dexterous Manipulation}
Natural human demonstrations often involve multiple objects and complex hand-object interactions, where the task goal, the manipulated object, and its relations to other objects in the scene are not explicitly specified. Therefore, task-level semantic parsing is required before hand-object state reconstruction. To this end, EgoAERO first employs an MLLM to perform lightweight semantic preprocessing on the raw RGB-D video. The system samples a small number of keyframes from the input video and feeds them, together with the task description, into the MLLM to identify the target manipulated object and its potentially related supporting objects, containers, or static objects. Based on the semantic parsing results, the MLLM generates text prompts for SAM3~\citep{kirillov2023segment,carion2025sam3} segmentation and selects a less-occluded keyframe as the initialization seed frame. SAM3 then uses these prompts to obtain frame-wise masks of the target object and related objects, which serve as inputs to the subsequent asset-free object tracking and reconstruction module. It is important to note that the MLLM is only used for semantic-level initialization and pipeline configuration; it does not directly estimate the object 6-DoF pose, object geometry, or hand pose.

\subsubsection{Asset-free Object Tracking and Reconstruction under Egocentric Hand Occlusion}
In egocentric manipulation scenarios without relying on object CAD models or pre-scanned meshes, object tracking and reconstruction must handle frequent hand occlusions, rapidly changing visible regions, and tracking drift caused by low-texture surfaces. Inspired by recent unknown-object RGB-D tracking and reconstruction systems~\citep{wen2021bundletrack,wen2023bundlesdf,wen2024foundationpose}, EgoAERO relies only on RGB-D observations and object masks, and recovers the 6-DoF pose trajectory and geometric representation of unknown objects through coarse pose initialization, keyframe memory-pool optimization, and coarse-to-fine mesh reconstruction. This provides stable object states for subsequent hand-object joint modeling.

\paragraph{Coarse Object Pose Initialization.}
To provide a reliable initialization for the subsequent keyframe-based pose optimization, EgoAERO first estimates a coarse object pose $\tilde{T}_t\in SE(3)$ for the current frame $F_t$. Given the RGB-D input and object mask $M_t^{O}$ at frame $t$, the system back-projects the visible object region into a local point cloud and initializes a canonical object frame $\mathcal{O}$ from the first reliable observation. EgoAERO then establishes local RGB-D correspondences between the visible object regions in the current frame $F_t$ and the previous frame $F_{t-1}$, and lifts the 2D matches to 3D correspondences using depth. To reduce the influence of outliers caused by hand occlusion, low-texture surfaces, and segmentation errors, the system adopts RANSAC~\citep{fischler1981ransac} for robust rigid pose estimation and selects the hypothesis with the highest inlier consistency as the coarse object pose $\tilde{T}_t$. This coarse pose is used only as the initialization for the subsequent memory-pool pose optimization, rather than as the final object pose.

\paragraph{Keyframe Memory-pool Pose Optimization.}
The coarse pose $\tilde{T}_t$ can still drift under hand occlusion, partial visibility, and low-texture observations. To stabilize tracking, EgoAERO maintains a keyframe memory pool $\mathcal{P}$ containing historical frames with reliable object observations and complementary viewpoints. For the current frame $F_t$, EgoAERO selects a small subset of relevant keyframes $\mathcal{K}_t\subset\mathcal{P}$ and constructs a local pose graph $\mathcal{G}_t=(\mathcal{V}_t,\mathcal{E}_t)$, where $\mathcal{V}_t=\{t\}\cup\mathcal{K}_t$. The current node is initialized by $\tilde{T}_t$, while memory nodes are initialized by their previously optimized poses. The object poses in this local graph are then jointly optimized by combining cross-frame correspondence, geometric, surface, silhouette, and pose-prior constraints:
\[
    \min_{\{T_i\}_{i\in\mathcal{V}_t}}
    \sum_{(i,j)\in\mathcal{E}_t}
    \left[
    \lambda_f E_{\mathrm{feat}}(i,j)
    +
    \lambda_g E_{\mathrm{geo}}(i,j)
    \right]
    +
    \sum_{i\in\mathcal{V}_t}
    \left[
    \lambda_s E_{\mathrm{sdf}}(i)
    +
    \lambda_m E_{\mathrm{mask}}(i)
    +
    \lambda_p E_{\mathrm{pose}}(i)
    \right]
\]
Here, $E_{\mathrm{feat}}$ and $E_{\mathrm{geo}}$ provide multi-view RGB-D alignment, $E_{\mathrm{sdf}}$ couples pose tracking with the online object geometry, $E_{\mathrm{mask}}$ constrains silhouette consistency, and $E_{\mathrm{pose}}$ prevents excessive deviation from the initial poses. After optimization, the current pose $T_t$ is used as the final tracking result and is written back to the memory pool when the frame provides reliable new view coverage. This local memory-pool optimization converts single-frame coarse tracking into a temporally stable 6-DoF object trajectory. Details of memory scoring, keyframe selection, graph construction, and residual definitions are provided in Appendix~\ref{app:keyframe_memory_pose}.

\paragraph{Neural Object Field Guided Coarse-to-Fine Mesh Reconstruction.}
Object observations in egocentric dexterous manipulation videos are usually incomplete. Directly reconstructing a high-fidelity mesh from local RGB-D observations can easily introduce noise and coordinate inconsistency, while relying only on an online neural field often provides temporally consistent but relatively coarse geometry. Therefore, EgoAERO adopts a coarse-to-fine object reconstruction strategy: it first uses an online neural object field to obtain an object-centric and temporally consistent coarse geometry, and then combines SAM3D~\citep{sam3dteam2025sam3d} with the original RGB-D observations to recover finer surface details. Specifically, it maintains an online neural object field $\Omega_{\Theta}$ in the object canonical frame $\mathcal{O}$ and fuses keyframe observations aligned by the optimized poses. The field is trained with occlusion-aware ray supervision:
\begin{equation*}
    \mathcal{L}_{\mathrm{obj}} =
    \lambda_{\mathrm{surf}}\mathcal{L}_{\mathrm{surf}}
    + \lambda_{\mathrm{free}}\mathcal{L}_{\mathrm{free}}
    + \lambda_{\mathrm{occ}}\mathcal{L}_{\mathrm{occ}}
    + \lambda_{\mathrm{rgb}}\mathcal{L}_{\mathrm{rgb}}
    + \lambda_{\mathrm{eik}}\mathcal{L}_{\mathrm{eik}}
\end{equation*}
This produces a temporally consistent coarse mesh $\mathcal{M}_{O}^{\mathrm{coarse}}$ from the zero level set of $\Omega_{\Theta}$. Second, EgoAERO uses the original RGB-D observations, object masks, and point maps as inputs to SAM3D to recover a more detailed mesh $\mathcal{M}_{O}^{\mathrm{sam}}$. Since $\mathcal{M}_{O}^{\mathrm{sam}}$ is not guaranteed to lie in the object canonical frame, EgoAERO aligns it to $\mathcal{M}_{O}^{\mathrm{coarse}}$ with rigid and scale alignment, yielding the final mesh $\mathcal{M}_{O}$. Thus, the final mesh inherits the coordinate consistency of the neural field while incorporating finer surface details from SAM3D. Details of ray supervision, loss terms, and mesh alignment are provided in Appendix~\ref{app:object_mesh_reconstruction}.

\subsubsection{Egocentric Hand Pose Estimation}

EgoAERO uses the camera-frame MANO~\citep{romero2017mano} results from HaWoR~\citep{zhang2025hawor} as the hand initialization, and further applies a lightweight correction to the global hand translation using RGB-D depth information. Specifically, given an RGB sequence, the hand estimator $\mathcal{H}_{\psi}$ outputs the MANO articulation, shape parameters, and hand root pose for each frame:
\begin{equation*}
    \left(
    \theta_t,\beta_t,{}^{C_t}T_{H_t}
    \right)_{t=1}^{T}
    =
    \mathcal{H}_{\psi}(I_{1:T}),
\end{equation*}
where $\theta_t$ denotes the MANO articulation, $\beta_t$ denotes the shape parameter, and ${}^{C_t}T_{H_t}\in SE(3)$ represents the rigid transformation from the hand root to the current camera frame $C_t$. To maintain shape consistency across the sequence, EgoAERO uses the sequence-level average shape $\bar{\beta}$ and obtains the hand mesh vertices and joints in the camera frame through MANO forward kinematics.

Since monocular hand pose estimation may suffer from global depth bias, EgoAERO further uses the RGB-D depth map to correct the global translation of the whole hand. Specifically, the system projects MANO vertices onto the RGB-D image, queries RGB-aligned depth values in local neighborhoods, and estimates a translation correction $\Delta p_t^C$ from the robust residuals between the visible 3D hand surface observations and the predicted vertices. This correction is applied only to the global hand translation. After this step, the hand mesh is more stably aligned with the RGB-D geometry, providing a reliable hand state for subsequent hand-object contact recovery.

\subsubsection{Ego Motion Compensation}

Since the egocentric camera is mounted on the head, head motion is mixed into the hand-object trajectories expressed in the camera frame: static objects may exhibit spurious drift, and the hand wrist trajectory may contain camera motion unrelated to manipulation. To address this issue, EgoAERO estimates the camera trajectory with an RGB-D SLAM backend~\citep{campos2021orbslam3} and transforms all frames into a fixed table frame $\mathcal{T}$. In this way, the hand-object states estimated in each camera frame are represented in a unified and stable coordinate frame. To reduce the influence of dynamic hand regions on SLAM, EgoAERO uses hand masks to down-weight hand pixels during camera trajectory estimation, allowing the background and tabletop regions to dominate ego motion recovery. After the transformation, EgoAERO applies only lightweight temporal smoothing to the object trajectory and hand root translation. It does not force the object to remain on the table or impose vertical constraints on the object’s local axes, avoiding incorrectly flattening real object rotations during grasping and manipulation.

\subsubsection{Adaptive Contact Optimization}
\label{sec:contact_optimization}

\begin{wrapfigure}[14]{r}{0.57\linewidth}
    \vspace{-0.6em}
    \centering
    \includegraphics[width=\linewidth]{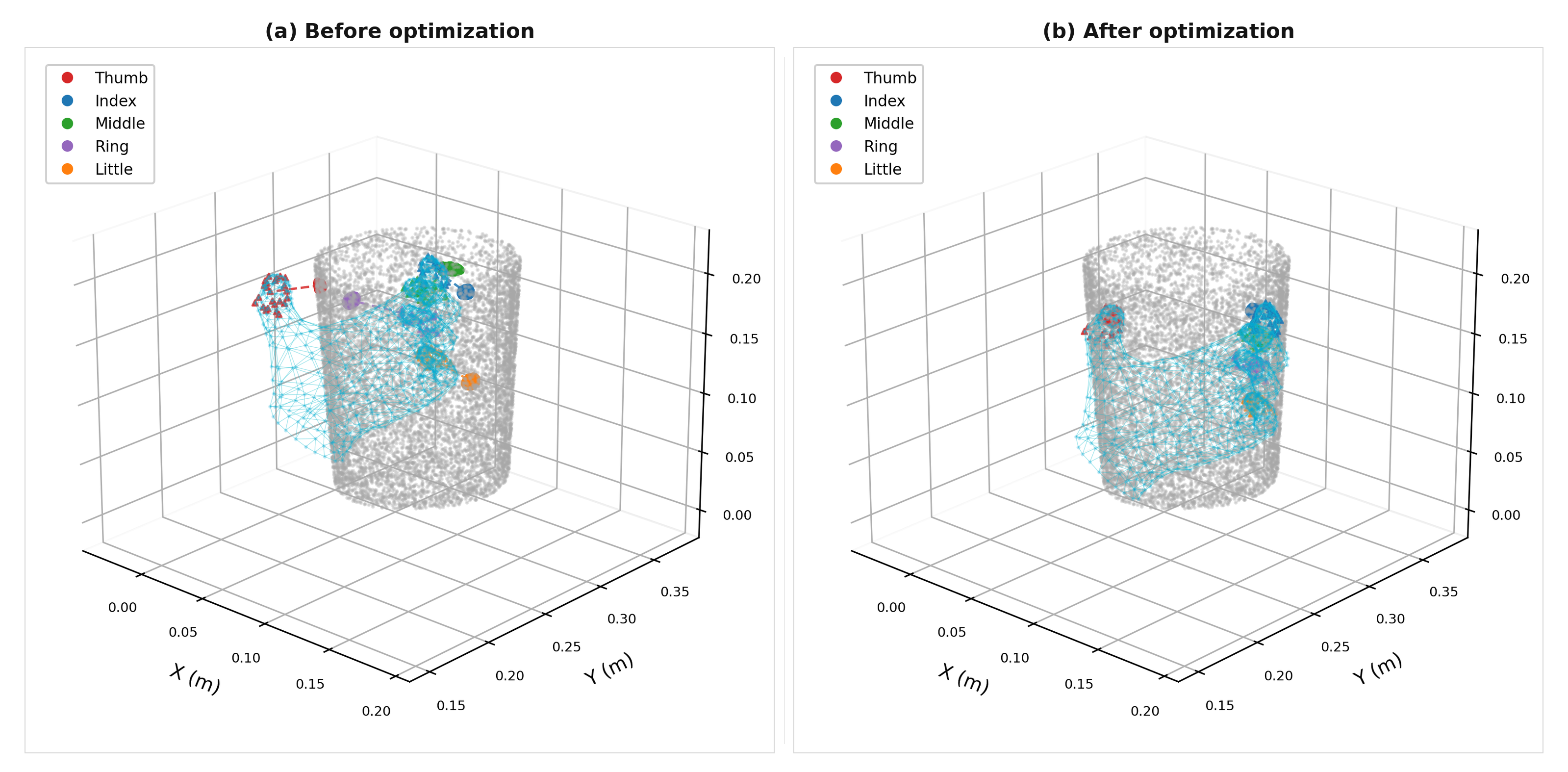}
    \vspace{-0.8em}
    \caption{
    \textbf{Adaptive contact optimization.}
    Before and after visualization of local hand--object contact correction.
    }
    \label{fig:contact_optimization}
    \vspace{-0.4em}
\end{wrapfigure}

Due to egocentric occlusions and hand pose estimation errors, fingertip floating, missing contacts, or local penetrations may still occur during grasping. EgoAERO formulates this problem as a conservative geometry-level contact correction: it keeps the object pose, object mesh, and MANO articulation unchanged, and only applies bounded corrections to the global hand translation and vertices in local contact regions. Specifically, EgoAERO first selects an active manipulation window based on the operation prior and hand-object distance, so that corrections are applied only around grasping, moving, and placing stages, avoiding erroneous attraction in non-manipulation periods. For each active frame, EgoAERO constructs three types of contact regions: the thumb pulp, a dynamically selected opposing non-thumb fingertip region, and the thenar region that helps stabilize grasping. The optimization first estimates a constrained whole-hand translation to correct the global hand-object misalignment, and then applies small local geometric offsets to the thumb and opposing fingers to enhance fingertip contact while keeping the palm and wrist stable. Finally, EgoAERO temporally smooths the correction trajectory, applies boundary tapering, and uses penetration push-back to suppress severe local penetrations. This module does not re-solve MANO parameters or modify the object trajectory and geometry, thereby improving the contact consistency and physical usability of the reconstructed data while preserving the original demonstration motion as much as possible. Detailed definitions of contact region selection, whole-hand translation, local finger correction, and penetration push-back are provided in Appendix~\ref{app:contact_optimization}.

\subsection{Two-stage Residual Policy Learning}
\label{sec:video_to_robot}

EgoAERO uses the reconstructed hand-object trajectories for policy learning, converting a single egocentric RGB-D demonstration into an executable dexterous robot manipulation policy. Following human-video-based dexterous imitation and two-stage residual learning ideas~\citep{qin2022dexmv,li2025maniptrans}, the first stage learns stable hand trajectory tracking, while the second stage learns small residual corrections under object and contact constraints. This decomposition avoids directly exploring high-dimensional dexterous hand actions from sparse task rewards.

\paragraph{Stage I: Hand Trajectory Tracking.}

The first-stage policy does not directly model object dynamics, but instead learns stable tracking of human hand motion with a dexterous robot hand. EgoAERO first reconstructs the human wrist pose and finger keypoint trajectories from a single ego demonstration, and uses them as the reference motion for robot hand tracking. Since the human hand and the dexterous robot hand have different morphologies, EgoAERO first obtains an initial robot hand trajectory through kinematic retargeting to warm-start policy training. However, this initialization only provides a reachable initial action sequence, while the supervision target remains the reconstructed human hand reference trajectory. Based on this, EgoAERO trains a hand tracking policy $\pi_I(a_t^I\mid s_t^I)$, so that the robot wrist and finger keypoints follow the human hand motion in the task frame while maintaining smooth actions. This stage does not require the object to be successfully manipulated; instead, it provides a stable hand control prior for the second-stage object-contact residual policy. Details of the reference trajectory definition and reward are provided in Appendix~\ref{app:policy_rewards}.

\paragraph{Stage II: Object-Contact Residual Policy Learning.}

Hand tracking alone is usually insufficient for stable object manipulation. Therefore, in the second stage, EgoAERO introduces object states and contact feedback on top of the hand tracking policy, and learns a residual policy $\pi_R$ to make small corrections to the base action $a_t^I$ produced by the hand tracking policy:
\[
    a_t = a_t^I + \Delta a_t^R,\qquad
    a_t^I\sim \pi_I(\cdot\mid s_t^I),\qquad
    \Delta a_t^R\sim \pi_R(\cdot\mid s_t^R,a_t^I)
\]
Here, $\Delta a_t^R$ denotes the residual correction. In addition to the hand tracking state, $s_t^R$ includes the object reference trajectory, the current object pose and velocity, the object geometry encoding, the hand-object distance, and simulated contact forces. The reward in the residual stage retains the hand imitation term and further incorporates object trajectory tracking and contact constraints, so that the robot can both stay close to the hand motion reconstructed by EgoAERO and drive the object to follow the reference trajectory with stable contact. In this way, EgoAERO converts the reconstructed hand-object trajectories into an executable closed-loop robot policy. Detailed reward definitions are provided in Appendix~\ref{app:policy_rewards}.

\section{Egocentric Demonstration Data Collection and Quality Assessment}
\label{sec:quality}

EgoAERO enables a closed-loop data collection pipeline for egocentric dexterous manipulation. During collection, the system reconstructs a provisional hand-object trajectory online, evaluates whether it is physically usable, and immediately decides whether to keep, repair, or recapture the demonstration.

\paragraph{Online Ego Data Quality Assessment.}
\begin{wraptable}{r}{0.52\textwidth}
    \vspace{-1.0em}
    \centering
    \caption{\scriptsize Ego dataset comparison.}
    \label{tab:dataset_comparison}
    \scriptsize
    \setlength{\tabcolsep}{3pt}
    \resizebox{\linewidth}{!}{
    \begin{tabular}{lcccccc}
        \toprule
        Dataset & Scale & Obj. state & Asset-free & Depth & SLAM & Contact eval. \\
        \midrule
        Ego4D & Large & \xmark & \cmark & \xmark & \xmark & \xmark \\
        EgoDex & Large & \xmark & \cmark & \xmark & \cmark & \xmark \\
        HOI4D & Medium & \cmark & \xmark & \cmark & \xmark & Partial \\
        EgoDex-R & Large & \cmark & \cmark & \cmark & \cmark & \cmark \\
        \bottomrule
    \end{tabular}}
    \vspace{-1.0em}
\end{wraptable}

The online assessment is based on \emph{bounded recoverability}: a sequence is considered useful if stable hand-object contact can be recovered through small, local, and interpretable corrections, without modifying the object trajectory or re-solving hand articulation. EgoAERO evaluates tracking stability, contact consistency, residual penetration, and temporal jitter from the reconstructed hand state, object pose, and object geometry. It outputs three collection decisions: \emph{accept}, \emph{repairable accept}, or \emph{recapture}. Complete scoring functions, thresholds, and per-finger diagnostics are provided in Appendix~\ref{app:online_quality_assessment}.

\paragraph{EgoDex-R: An Egocentric Dataset for Dexterous Manipulation Learning.}

 Using this collection loop, we construct EgoDex-R, an egocentric dexterous manipulation dataset collected with FastUMI Ego and without object CAD assets. EgoDex-R contains approximately 4.3M RGB-D frames, 5,600 manipulation sequences, and over 1,000 target object instances across more than 200 daily task categories. As shown in Table~\ref{tab:dataset_comparison}, EgoDex-R is distinguished by providing contact-consistent hand-object trajectories with object pose, object geometry, depth, SLAM, and contact quality annotations. These structured outputs make the dataset directly usable for robot retargeting, imitation learning, residual policy learning, and simulation replay. Detailed data fields and collection statistics are provided in Appendix~\ref{app:egodexr_details}.
\par\wrapfill

\section{Experiments}

We evaluate EgoAERO from two perspectives: (1) Can a single ego RGB-D demonstration reconstructed by EgoAERO drive a robot to accomplish dexterous manipulation tasks? We evaluate this through both simulation and real-world experiments. (2) Without using object asset priors, can the trajectories reconstructed by EgoAERO achieve performance comparable to data with object priors?

\subsection{Simulation Experiment Setting}
All simulation experiments are conducted in Isaac Gym~\citep{makoviychuk2021isaac} with the two-stage residual policy learning pipeline in Sec.~\ref{sec:video_to_robot}. For each ego demonstration, EgoAERO reconstructs an object mesh, an object reference trajectory, and a human hand motion trajectory, which are then converted into a dexterous manipulation task in simulation. We evaluate two settings. First, on EgoDex-R, we test whether a single ego demonstration can drive manipulation, and compare the full EgoAERO pipeline with two ablations: \textit{Only Hand Pose} and \textit{w/o Adaptive Contact Optimization}. Second, on HOI4D, we compare EgoAERO's asset-free reconstruction from raw RGB-D input with trajectories reconstructed using object CAD assets. We report object rotation error ($\mathrm{E_r}$), object translation error ($\mathrm{E_t}$), mean joint position error ($\mathrm{E_j}$), mean fingertip position error ($\mathrm{E_{ft}}$), and success rate ($\mathrm{SR}$). Detailed protocols and metric definitions are provided in Appendix~\ref{app:experiment_protocol} and Appendix~\ref{app:experiment_metrics}.

\subsection{Simulation Experiment Results}

\begin{table}[t]
    \centering
    \caption{Dexterous manipulation results. The first four metrics are averaged over successful rollouts.}
    \vspace{-0.1em}
    \label{tab:main_results}
    \setlength{\tabcolsep}{4pt}
    \resizebox{0.92\textwidth}{!}{
    \begin{tabular}{llccccc}
        \toprule
        Dataset & Method & $\mathrm{E_r}\downarrow$ & $\mathrm{E_t}\downarrow$ & $\mathrm{E_j}\downarrow$ & $\mathrm{E_{ft}}\downarrow$ & $\mathrm{SR}(\%)\uparrow$ \\
        \midrule
        EgoDex-R & \textit{Only Hand Pose} & 28.6 & 4.72 & 3.35 & 2.48 & 9.8 \\
        EgoDex-R & \textit{w/o Adaptive Contact Optimization} & 15.4 & 1.36 & 2.93 & 2.18 & 36.2 \\
        EgoDex-R & \textbf{EgoAERO} & \textbf{9.7} & \textbf{0.82} & \textbf{2.48} & \textbf{1.65} & \textbf{49.5} \\
        \midrule
        HOI4D & \textit{Raw Data (with Object CAD)} & \textbf{10.4} & 0.73 & \textbf{2.32} & 1.69 & 43.3 \\
        HOI4D & \textbf{EgoAERO} & 10.9 & \textbf{0.68} & 2.44 & \textbf{1.58} & \textbf{44.7} \\
        \bottomrule
    \end{tabular}
    }
\end{table}
Table~\ref{tab:main_results} reports the simulation results. On EgoDex-R, using only hand pose leads to low success rates, showing that hand motion alone is insufficient for dexterous object manipulation without object state and contact information. Introducing asset-free object reconstruction greatly reduces object tracking errors and improves task success, while adaptive contact optimization further improves hand tracking quality and contact stability.

On HOI4D, EgoAERO achieves performance close to the raw CAD-based annotations and performs better on some metrics. One possible reason is that the raw annotations are mainly designed for perception-level reconstruction, whereas EgoAERO further applies adaptive contact optimization before policy learning. This step improves the physical consistency of the hand-object demonstrations by reducing fingertip floating, missing contact, and local penetration. These results indicate that, even without object CAD assets, EgoAERO can reconstruct hand-object demonstrations with consistent object motion and physically plausible contact, providing effective supervision for downstream policy learning.

\begin{figure}
    \centering
    \includegraphics[width=1\linewidth]{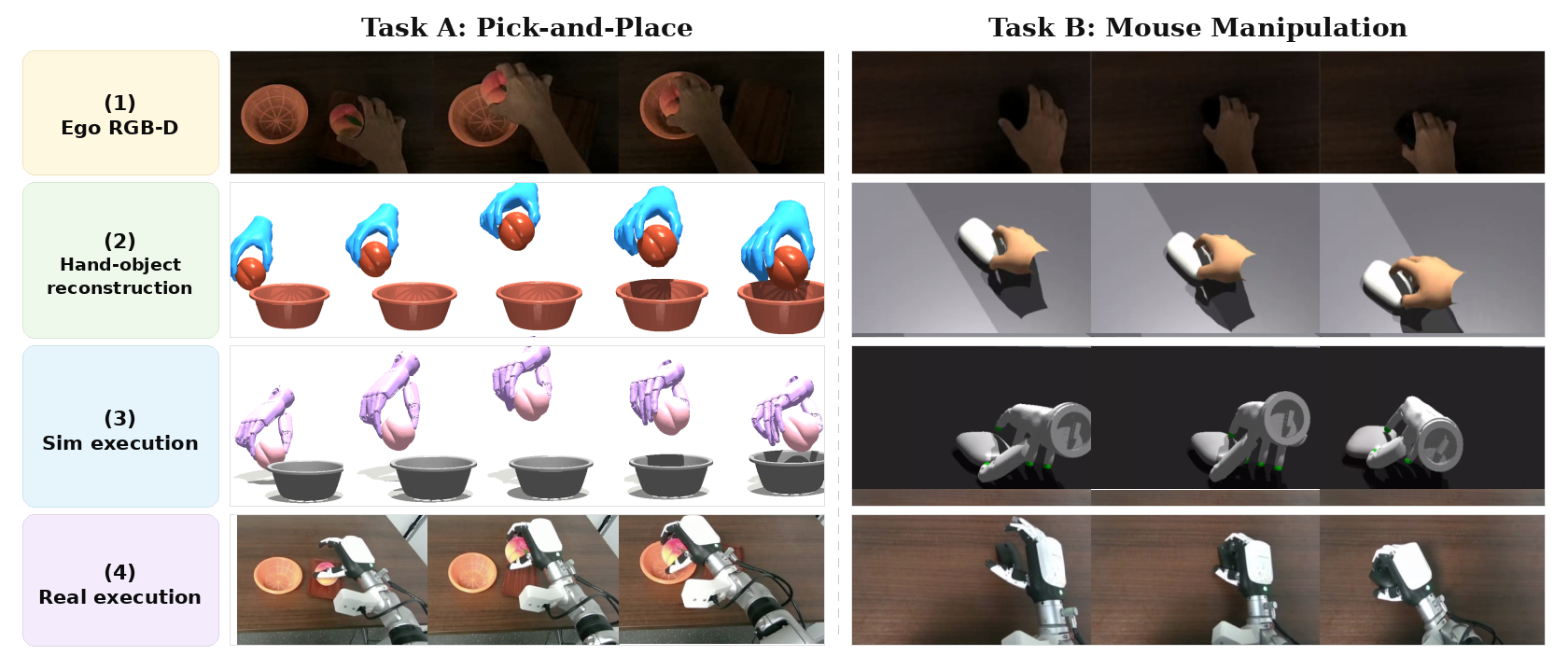}
    \vspace{-0.7em}
    \caption{Qualitative demonstration of EgoAERO. 
From a single egocentric human video, EgoAERO reconstructs the hand-object trajectory, transfers it to a simulated dexterous hand, and enables real-world robot execution. 
}
    \label{fig:real_world_demo}
\end{figure}

\subsection{Real-World Experiment}

We further evaluate EgoAERO on a real robot platform composed of a Unitree G1 humanoid robot and an Inspire Hand, as shown in Fig.~\ref{fig:real_world_demo}. For each task, EgoAERO first reconstructs the hand-object trajectory from a single egocentric demonstration, and then trains the two-stage residual policy described in Sec.~\ref{sec:video_to_robot}. The learned policy generates the dexterous hand and wrist trajectory used for execution. On hardware, the G1 arm tracks the policy-generated wrist motion, while the Inspire Hand executes the policy-generated finger commands. During real-world execution, we do not enforce strict temporal synchronization with the original human demonstration, since the physical robot may move more slowly than the human hand. These experiments verify that EgoAERO can transform a natural ego demonstration into a contact-consistent robot trajectory that can be executed on physical dexterous hardware.

\section{Conclusion}

\label{sec:conclusion}

We presented EgoAERO, an asset-free framework that converts a single egocentric RGB-D human demonstration into structured hand-object trajectories and executable dexterous manipulation policies. EgoAERO combines MLLM-guided semantic initialization, asset-free object tracking and reconstruction, egocentric hand pose estimation, ego-motion compensation, and adaptive contact optimization to recover replayable hand-object demonstrations without requiring object CAD models or pre-scanned meshes. Based on these reconstructed trajectories, a two-stage residual policy learning pipeline transfers natural human demonstrations to dexterous robot execution. Experiments on EgoDex-R and HOI4D show that EgoAERO substantially improves over hand-only imitation, benefits from contact optimization, and achieves performance close to CAD-based trajectories while avoiding object asset priors. Real-world experiments further demonstrate that the reconstructed trajectories can be transferred to physical dexterous hardware.

\section{Limitations}

\label{sec:limitations}

EgoAERO still has several limitations. Its current pipeline mainly targets single-hand manipulation, and performance can degrade under severe occlusion, reflective objects, fast motion, or inaccurate hand/segmentation estimates. In addition, policy learning is still task-specific and requires simulation training, leaving cross-task generalization and lower training cost as future directions.

\clearpage
\bibliography{main}  

\clearpage
\appendix
\section{Details of Keyframe Memory-pool Pose Optimization}
\label{app:keyframe_memory_pose}

\paragraph{Memory-frame representation and quality score.}
EgoAERO maintains a keyframe memory pool $\mathcal{P}$ to store informative historical observations for local pose graph optimization. Each memory frame is represented as
\[
F_k=\{I_k,D_k,M_k^{O},M_k^{H},P_k^{C},T_k,q_k\},
\]
where $I_k$ and $D_k$ denote the RGB image and depth map, $M_k^{O}$ and $M_k^{H}$ denote the object and hand masks, $P_k^{C}$ is the local object point cloud in the camera frame, $T_k$ is the optimized object pose, and $q_k$ is the observation quality score.

For a new frame $F_t$, EgoAERO computes its quality score as
\[
q_t=
\alpha_v A_t^{O}
+\alpha_d R_t^{D}
+\alpha_\theta C_t^{\theta}
-\alpha_h H_t^{\mathrm{occ}},
\]
where $A_t^{O}$ measures the visible object area, $R_t^{D}$ measures the valid depth ratio within the object region, $C_t^{\theta}$ measures the view complementarity with respect to existing memory frames, and $H_t^{\mathrm{occ}}$ measures the hand occlusion level. A frame is inserted into $\mathcal{P}$ only when $q_t$ is above a threshold and the frame provides additional view coverage.

\paragraph{Keyframe subset selection.}
Given the current frame $F_t$, EgoAERO selects a subset of memory frames $\mathcal{K}_t\subset\mathcal{P}$ for local pose graph optimization. For each candidate memory frame $F_k$, the selection score is defined as
\[
s(k,t)=
\beta_o \operatorname{Overlap}(F_k,F_t)
-
\beta_r d_R(R_k,\tilde{R}_t)
+
\beta_q q_k ,
\]
where $\operatorname{Overlap}(F_k,F_t)$ measures the visible-region overlap between the candidate frame and the current frame, $d_R(R_k,\tilde{R}_t)$ is the geodesic distance between their object rotations, and $q_k$ is the quality score of the candidate memory frame. Here, $R_k$ and $\tilde{R}_t$ are the rotation components of the optimized pose $T_k$ and the coarse pose $\tilde{T}_t$, respectively. EgoAERO selects the top-$K$ memory frames according to $s(k,t)$ and constructs a local pose graph
\[
\mathcal{G}_t=(\mathcal{V}_t,\mathcal{E}_t),
\qquad
\mathcal{V}_t=\{t\}\cup\mathcal{K}_t .
\]

\paragraph{Pose graph objective.}
In the local pose graph, the current frame is initialized with the coarse pose $\tilde{T}_t$, while memory frames are initialized with their previously optimized poses. EgoAERO jointly optimizes the object poses of all nodes:
\[
    \min_{\{T_i\}_{i\in\mathcal{V}_t}}
    \sum_{(i,j)\in\mathcal{E}_t}
    \left[
    \lambda_f E_{\mathrm{feat}}(i,j)
    +
    \lambda_g E_{\mathrm{geo}}(i,j)
    \right]
    +
    \sum_{i\in\mathcal{V}_t}
    \left[
    \lambda_s E_{\mathrm{sdf}}(i)
    +
    \lambda_m E_{\mathrm{mask}}(i)
    +
    \lambda_p E_{\mathrm{pose}}(i)
    \right].
\]
The feature term $E_{\mathrm{feat}}$ measures the consistency of cross-frame RGB-D correspondences after transforming them into the object canonical frame. Given the correspondence set $\mathcal{C}_{ij}$ between frame $i$ and frame $j$, it is written as
\[
E_{\mathrm{feat}}(i,j)
=
\sum_{(p_i,p_j)\in\mathcal{C}_{ij}}
\rho
\left(
\left\|
T_i^{-1}p_i - T_j^{-1}p_j
\right\|_2
\right),
\]
where $p_i$ and $p_j$ are 3D correspondence points in their camera frames, and $\rho(\cdot)$ is a robust kernel.

The geometric term $E_{\mathrm{geo}}$ enforces point-to-plane consistency between local depth observations. For a point $p$ in frame $i$, EgoAERO transforms it into frame $j$ using the current pose estimates, finds its projective association in the depth map of frame $j$, and penalizes the point-to-plane distance:
\[
E_{\mathrm{geo}}(i,j)
=
\sum_{p\in P_i^C}
\rho
\left(
\left|
n_j(\hat{p})
^\top
\left(
\hat{p}-T_{j}^{-1}T_i p
\right)
\right|
\right),
\]
where $\hat{p}$ is the associated 3D point in frame $j$, and $n_j(\hat{p})$ is the corresponding surface normal.

The SDF term $E_{\mathrm{sdf}}$ attracts aligned observations to the current implicit object surface:
\[
E_{\mathrm{sdf}}(i)
=
\sum_{p\in P_i^C}
\rho
\left(
\left|
\Omega_{\Theta}(T_i^{-1}p)
\right|
\right),
\]
where $\Omega_{\Theta}$ denotes the online neural object field defined in the object canonical frame. The mask term $E_{\mathrm{mask}}$ imposes weak silhouette consistency between the projected object model and the observed object mask, while the pose regularization term prevents the solution from drifting too far from the initialization:
\[
E_{\mathrm{pose}}(i)
=
\left\|
\log(\bar{T}_i^{-1}T_i)
\right\|_2^2 ,
\qquad
\bar{T}_i=
\begin{cases}
\tilde{T}_t, & i=t,\\
T_i^{\mathrm{old}}, & i\in\mathcal{K}_t .
\end{cases}
\]
Here, $\bar{T}_i$ denotes the initialization pose of node $i$.

\paragraph{Pose update and memory update.}
The pose variables are updated in the Lie algebra:
\[
T_i \leftarrow \exp(\delta \xi_i^\wedge)T_i,
\qquad
\delta \xi_i\in\mathfrak{se}(3).
\]
After optimization, the optimized current pose $T_t$ is used as the final tracking result. If the current frame satisfies the memory update condition, it is inserted into $\mathcal{P}$ with its optimized pose and quality score. Selected historical memory frames can also be updated with their refined poses, which helps reduce early tracking errors and improves the consistency of subsequent geometry fusion.

\section{Details of Coarse-to-Fine Object Mesh Reconstruction}
\label{app:object_mesh_reconstruction}

This appendix provides the details of the object mesh reconstruction module. EgoAERO defines an online neural object field $\Omega_{\Theta}$ in the object canonical frame $\mathcal{O}$. Given optimized keyframe poses, RGB-D observations are transformed into $\mathcal{O}$ and used to sample points along depth rays. For a sampled point $x\in\mathbb{R}^3$, the geometry network predicts the SDF value $s_x=\Omega_{\Theta}(x)$.

\section{Details of Adaptive Contact Optimization}
\label{app:contact_optimization}

This appendix provides the implementation details of the adaptive contact optimization described in the main paper. The inputs to the optimization are the hand vertices $V_t^{\mathcal{T}}$, hand joints $J_t^{\mathcal{T}}$, the object mesh $\mathcal{M}_{\mathcal{O}}$, and the object pose ${}^{\mathcal{T}}T_{\mathcal{O},t}=(R_t,q_t)$ in the table frame $\mathcal{T}$. After sampling surface points and normals $\{(p_i,n_i)\}_{i=1}^{N_s}$ from the object mesh in its local frame, the object surface at frame $t$ can be written as
\begin{equation}
    \hat{p}_{i,t}=R_t p_i+q_t,\qquad
    \hat{n}_{i,t}=R_t n_i .
\end{equation}
The optimization keeps the object pose, object mesh, MANO shape, and original articulation unchanged, and only updates the replay hand geometry:
\begin{equation}
    V_t^{\mathcal{T}}\rightarrow V_t^{\prime\mathcal{T}},\qquad
    J_t^{\mathcal{T}}\rightarrow J_t^{\prime\mathcal{T}} .
\end{equation}

\paragraph{Active window and contact regions.}
The optimization is applied only to the active frame set $\mathcal{W}$. In practice, $\mathcal{W}$ is determined by hand validity, script-specified frame ranges, and an optional operation prior. If semantic stage priors are available, only the grasp, move, and place stages, together with their padded ranges, are retained. For each active frame, EgoAERO constructs three candidate contact regions: the thumb pulp $\mathcal{C}_t^{thumb}$, the opposing non-thumb fingertip region $\mathcal{C}_t^{opp}$, and the thenar region $\mathcal{C}_t^{hukou}$. The opposing finger is not fixed, but dynamically selected frame by frame from the index, middle, ring, and little fingers according to the closest distance to the object:
\begin{equation}
    f_t^{opp}
    =
    \arg\min_{f\in\{\mathrm{index,middle,ring,little}\}}
    D_f(t),
\end{equation}
where $D_f(t)$ denotes the nearest-neighbor distance statistic between the fingertip vertices of finger $f$ and the object surface. This strategy allows the optimization to adapt to different contact patterns in two-finger, three-finger, and multi-finger grasps.

\paragraph{Global hand translation.}
Let $\pi_t(x)$ denote the nearest point of a hand point $x$ on the current object surface, and let $n_{\pi_t(x)}$ be the corresponding surface normal. The signed distance is approximated as
\begin{equation}
    s_t(x)=(x-\pi_t(x))^\top n_{\pi_t(x)} .
\end{equation}
If a contact point is still floating relative to the target contact gap $g_k$, it is pulled closer along the object normal. For a contact region $\mathcal{C}_t^k$, the attraction direction is defined as
\begin{equation}
    d_t^k
    =
    \frac{1}{|\mathcal{C}_t^k|}
    \sum_{x\in\mathcal{C}_t^k}
    -n_{\pi_t(x)}
    \operatorname{ReLU}\left(s_t(x)-g_k\right),
    \qquad
    k\in\{\mathrm{thumb,opp,hukou}\}.
\end{equation}
The three regions are aggregated with region weights to obtain the whole-hand translation, which is then clipped by the maximum correction magnitude:
\begin{equation}
    \Delta_t^{raw}
    =
    \frac{\sum_k w_k d_t^k}{\sum_k w_k},
    \qquad
    \Delta_t^{global}
    =
    \operatorname{clip}
    \left(\Delta_t^{raw},\Delta_{\max}\right).
\end{equation}
By default, EgoAERO does not apply whole-hand rotation and only uses whole-hand translation to correct the global relative misalignment between the hand and the object.

\paragraph{Temporal smoothing and local finger correction.}
To avoid jitter caused by frame-wise nearest-neighbor queries, the whole-hand translation is smoothed with a finite-window triangular kernel and multiplied by taper weights at the boundaries of active segments:
\begin{equation}
    \tilde{\Delta}_t^{global}
    =
    b_t
    \sum_{\tau\in\mathcal{N}(t)}
    K(t-\tau)\Delta_\tau^{global},
\end{equation}
where $K$ is the triangular kernel and $b_t$ is the boundary transition weight. After applying the whole-hand translation, EgoAERO recomputes the contact distances of the thumb and the opposing finger, and estimates small local geometric offsets $\delta_t^{thumb}$ and $\delta_t^{opp}$. For a vertex of finger $f$, the update is defined as
\begin{equation}
    v_i'
    =
    v_i
    +
    \tilde{\Delta}_t^{global}
    +
    \alpha_i^f\delta_t^f ,
\end{equation}
where $\alpha_i^f$ is determined by the MANO finger chain. Vertices closer to the distal fingertip receive larger weights, while palm and wrist vertices have weights close to zero. The joints are updated in the same way using the corresponding finger-chain weights. This local correction does not re-solve the MANO pose or beta, but only enhances fingertip contact at the replay-geometry level.

\paragraph{Penetration rollback.}
After contact attraction, EgoAERO performs signed-distance checks on hand points near the object. If obvious deep penetration occurs, the penetrating point set is defined as
\begin{equation}
    \mathcal{P}_t
    =
    \{x\mid s_t(x)<-\epsilon\},
    \qquad
    depth(x)=\operatorname{ReLU}(-\epsilon-s_t(x)).
\end{equation}
A global push-back vector is then estimated along the object normals:
\begin{equation}
    r_t
    =
    \operatorname{clip}
    \left(
    \frac{1}{|\mathcal{P}_t|}
    \sum_{x\in\mathcal{P}_t}
    depth(x)n_{\pi_t(x)},
    r_{\max}
    \right),
\end{equation}
and applied to the hand vertices and joints. In implementation, EgoAERO allows light contact or small local penetration, but suppresses obvious deep penetration. The optimized hand geometry is written back to both the table frame and the camera frame for replay, retargeting package construction, and subsequent error analysis.

By default, EgoAERO uses a contact gap of $0.5$ mm, a thenar gap of $1.8$ mm, a maximum whole-hand translation of $34$ mm, a maximum local finger displacement of $15$ mm, and a maximum penetration push-back of $8$ mm. The upper bound for whole-hand rotation is set to $0^\circ$, meaning that whole-hand rotation is disabled by default. Both the whole-hand translation and local finger corrections use a temporal smoothing window of length $9$, and a $6$-frame boundary transition is applied at the beginning and end of each active segment.

\section{Reward Definitions for Two-stage Policy Learning}
\label{app:policy_rewards}

The first-stage objective is to track the human hand reference trajectory $\tau_h^H$ reconstructed by EgoAERO. The retargeted robot hand trajectory $\tau_h^{\mathrm{init}}$ is only used to initialize policy training, rather than serving as the direct tracking target in the reward. The hand tracking reward is defined as
\[
    r_t^{I}
    =
    w_w r_t^{\mathrm{wrist}}
    +
    w_f r_t^{\mathrm{finger}}
    +
    w_s r_t^{\mathrm{smooth}} .
\]
The wrist tracking reward uses the reconstructed human wrist trajectory as the reference and constrains position, orientation, and velocity:
\[
    r_t^{\mathrm{wrist}}
    =
    \exp
    \left(
    -\lambda_p\|p_{W,t}-p_{W,t}^{H}\|_2^2
    -\lambda_R d_R(R_{W,t},R_{W,t}^{H})^2
    -\lambda_v\|\dot{p}_{W,t}-\dot{p}_{W,t}^{H}\|_2^2
    \right).
\]
The finger imitation reward encourages the robot fingertips and finger keypoints to track the corresponding human hand keypoints:
\[
    r_t^{\mathrm{finger}}
    =
    \frac{1}{|\mathcal{K}_h|}
    \sum_{k\in\mathcal{K}_h}
    \exp
    \left(
    -\lambda_k
    \|x_{k,t}-x_{k,t}^{H}\|_2^2
    \right),
\]
where $\mathcal{K}_h$ denotes the set of hand keypoints with established correspondence after retargeting, including fingertips and finger keypoints. The smoothness reward suppresses high-frequency actions and unnecessary energy consumption, where $\tau_t$ denotes the actuator torque:
\[
    r_t^{\mathrm{smooth}}
    =
    \exp
    \left(
    -\lambda_a\|a_t^I-a_{t-1}^I\|_2^2
    -\lambda_\tau\|\tau_t\odot \dot{q}_t\|_1
    \right).
\]

The second stage trains a residual policy by adding object and contact constraints on top of the hand trajectory tracking reward. The reward is defined as
\[
    r_t^{R}
    =
    \eta_I r_t^I
    +
    \eta_o r_t^{\mathrm{obj}}
    +
    \eta_c r_t^{\mathrm{contact}}
    +
    \eta_\Delta r_t^{\mathrm{res}} .
\]
Here, the second-stage $r_t^I$ is computed from the robot state after executing the residual action, so as to prevent the residual policy from breaking the original hand motion structure. The object tracking reward encourages the simulated object to follow the reference object trajectory reconstructed by EgoAERO:
\[
    r_t^{\mathrm{obj}}
    =
    \exp
    \left(
    -\mu_p\|p_{O,t}-p_{O,t}^{\mathrm{ref}}\|_2^2
    -\mu_R d_R(R_{O,t},R_{O,t}^{\mathrm{ref}})^2
    -\mu_v\|\dot{p}_{O,t}-\dot{p}_{O,t}^{\mathrm{ref}}\|_2^2
    \right).
\]
The contact reward is applied only to fingers that are expected to be in contact according to the reference trajectory. Let $\mathcal{A}_t^{\mathrm{ref}}$ denote the reference active contact finger set at frame $t$, $d_{f,t}$ denote the distance from robot finger $f$ to the object surface, and $F_{f,t}$ denote the corresponding contact force. The contact reward is
\[
    r_t^{\mathrm{contact}}
    =
    \frac{1}{|\mathcal{A}_t^{\mathrm{ref}}|}
    \sum_{f\in\mathcal{A}_t^{\mathrm{ref}}}
    \exp(-\mu_d d_{f,t}^2)
    \left(
    1-\exp(-\mu_F\|F_{f,t}\|_2)
    \right).
\]
When $\mathcal{A}_t^{\mathrm{ref}}$ is empty, this contact term is skipped for the current frame. The residual regularization limits the second-stage policy from deviating excessively from the first-stage hand prior:
\[
    r_t^{\mathrm{res}}
    =
    \exp
    \left(
    -\mu_\Delta\|\Delta a_t^R\|_2^2
    \right).
\]
During training, an episode is terminated early if the object pose error, hand tracking error, or non-contact penetration exceeds a predefined threshold, which improves the sampling efficiency of residual policy learning.

\section{Details of Online Ego Data Quality Assessment}
\label{app:online_quality_assessment}

This appendix provides the detailed definition of the online quality assessment module. Given the coarse trajectory produced by EgoAERO in real time,
\begin{equation}
    x_0=\{H_t,O_t,\mathcal{M}_{\mathcal{O}}\}_{t=1}^{T},
\end{equation}
where $H_t$ denotes the hand state, $O_t$ denotes the object pose, and $\mathcal{M}_{\mathcal{O}}$ denotes the reconstructed object geometry, the quality assessment module solves a constrained projection:
\begin{equation}
    x^\star=
    \arg\min_x
    E_{\mathrm{contact}}(x)
    +E_{\mathrm{pen}}(x)
    +E_{\mathrm{temp}}(x)
    +\lambda\|x-x_0\|^2 ,
    \label{eq:bounded_recoverability}
\end{equation}
with bounded corrections:
\begin{equation}
    \|\Delta H_t^f\|\leq \delta_{\max},\qquad
    O_t=O_t^0 .
\end{equation}
This constraint allows the system to repair small errors around finger contact regions, but prevents it from producing high-quality results by significantly moving the wrist, modifying the object trajectory, or changing the hand articulation.

For each candidate finger $f$ in the active manipulation window $\mathcal{W}$, let $\mathcal{P}_f$ denote its fingertip pad vertex set. The module first computes the contact gap before and after the bounded projection:
\begin{equation}
    g_{t}^{f,\mathrm{before}}
    =
    \operatorname{median}_{v\in\mathcal{P}_f}
    d(v_t,\mathcal{M}_{\mathcal{O}}),
    \qquad
    g_{t}^{f,\mathrm{after}}
    =
    \operatorname{median}_{v\in\mathcal{P}_f}
    d(v_t^\star,\mathcal{M}_{\mathcal{O}}),
\end{equation}
where $d(\cdot,\mathcal{M}_{\mathcal{O}})$ denotes the distance to the object surface. The contact recoverability of finger $f$ is defined as
\begin{equation}
    Q_{\mathrm{rec}}^{f}
    =
    \frac{1}{|\mathcal{W}|}
    \sum_{t\in\mathcal{W}}
    \mathbf{1}
    \left[
    g_{t}^{f,\mathrm{after}}<\epsilon_g
    \land
    \|\Delta H_t^f\|<\epsilon_{\Delta}
    \right].
\end{equation}
This metric measures whether a finger can recover stable and plausible contact within a limited correction budget, rather than simply checking whether contact is achieved after optimization.

To prevent over-repair, the system records the repair budget usage
\begin{equation}
    B_{\mathrm{repair}}
    =
    \frac{
    \operatorname{median}_{t,f}\|\Delta H_t^f\|
    }{\delta_{\max}},
\end{equation}
as well as residual penetration, residual contact gaps, and the ratio of object motion without recoverable contact. The final quality score is
\begin{equation}
    Q=
    \exp
    \left(
    -\alpha R_{\mathrm{after}}
    -\beta B_{\mathrm{repair}}
    -\gamma U_{\mathrm{unresolved}}
    \right),
\end{equation}
where $R_{\mathrm{after}}$ denotes the remaining contact and penetration residuals after repair, and $U_{\mathrm{unresolved}}$ denotes failure modes that cannot be explained by local correction, such as severe object tracking failure or erroneous hand articulation. In practice, the system also outputs per-finger contact states, failure attribution, repair budget usage, and visualization reports, which support three collection decisions: \emph{accept}, \emph{repairable accept}, and \emph{recapture}.

\section{EgoDex-R Dataset Details}
\label{app:egodexr_details}

Each EgoDex-R sequence contains synchronized raw observations, reconstructed hand-object states, and task-level metadata. The raw observations include the egocentric RGB video, aligned depth maps, camera intrinsics, and timestamps. EgoAERO further provides SLAM camera poses in the table frame, MANO hand pose and shape parameters, hand mesh vertices and joints, target object 6-DoF pose trajectories, reconstructed object meshes, object masks, contact windows, and per-frame quality diagnostics.

In addition to geometric annotations, each sequence contains a task description generated or verified from the collection protocol, including the manipulated object, the intended action, and relevant relational objects when present. We also assign a difficulty score from $1$ to $5$ using an MLLM-based evaluator. The score considers interaction complexity, hand-object occlusion level, object motion difficulty, contact richness, and the expected difficulty of policy learning. These metadata fields allow EgoDex-R to support filtering, curriculum construction, failure analysis, and task-conditioned policy training.

\section{Simulation Experiment Protocol}
\label{app:experiment_protocol}

\paragraph{Datasets and comparisons.}
For EgoDex-R, we randomly select 100 task sequences to evaluate whether a single egocentric RGB-D demonstration can drive dexterous manipulation. We compare the full EgoAERO pipeline with two ablations: \textit{Only Hand Pose}, which uses hand motion without reconstructed object geometry and object pose supervision, and \textit{w/o Adaptive Contact Optimization}, which disables the contact refinement module before policy learning. For HOI4D, we randomly select 100 sequences and compare two sources of demonstration trajectories: \textit{Raw Data (with Object CAD)}, which uses the available object assets and annotations, and EgoAERO, which reconstructs hand-object trajectories directly from raw RGB-D videos without object assets.

\paragraph{Policy learning and evaluation.}
All methods are evaluated in the same Isaac Gym environment with the same dexterous hand model, reward terms, policy architecture, and training budget. Each reconstructed demonstration is converted into a simulation task consisting of an object mesh, an object reference trajectory, and a hand reference trajectory. We train policies using the two-stage procedure in Sec.~\ref{sec:video_to_robot}. At test time, each task is evaluated with multiple rollout seeds; success rate is computed over all evaluated rollouts. The object and hand tracking errors in Table~\ref{tab:main_results} are averaged over successful rollouts, while failed rollouts are reflected by the success rate.

\section{Definitions of Evaluation Metrics}
\label{app:experiment_metrics}

This appendix provides the definitions of the evaluation metrics used in the simulation experiments. Following ManipTrans~\citep{li2025maniptrans}, we evaluate policy performance from three aspects: object trajectory tracking, hand motion imitation, and task success.

\paragraph{Object rotation error.}
The average object rotation error $\mathrm{E_r}$ measures the mean rotational deviation between the simulated object pose and the reference object pose:
\[
    \mathrm{E_r}
    =
    \frac{1}{T}
    \sum_{t=1}^{T}
    d_R
    \left(
    R_{O,t},
    R_{O,t}^{\mathrm{ref}}
    \right),
\]
where $R_{O,t}$ is the object rotation in simulation at frame $t$, $R_{O,t}^{\mathrm{ref}}$ is the reference object rotation reconstructed by EgoAERO, and $d_R(\cdot,\cdot)$ denotes the geodesic distance between two rotations. $\mathrm{E_r}$ is reported in degrees.

\paragraph{Object translation error.}
The average object translation error $\mathrm{E_t}$ measures the mean positional deviation between the simulated object and the reference object:
\[
    \mathrm{E_t}
    =
    \frac{1}{T}
    \sum_{t=1}^{T}
    \left\|
    p_{O,t}
    -
    p_{O,t}^{\mathrm{ref}}
    \right\|_2,
\]
where $p_{O,t}$ is the simulated object position and $p_{O,t}^{\mathrm{ref}}$ is the reference object position. $\mathrm{E_t}$ is reported in centimeters.

\paragraph{Mean joint position error.}
The mean joint position error $\mathrm{E_j}$ measures the average joint-level tracking error between the robot hand and the reconstructed human hand:
\[
    \mathrm{E_j}
    =
    \frac{1}{T|\mathcal{J}|}
    \sum_{t=1}^{T}
    \sum_{j\in\mathcal{J}}
    \left\|
    x_{j,t}^{R}
    -
    x_{j,t}^{H}
    \right\|_2,
\]
where $\mathcal{J}$ is the set of evaluated hand joints, $x_{j,t}^{R}$ denotes the position of the $j$-th robot hand joint at frame $t$, and $x_{j,t}^{H}$ denotes the corresponding reconstructed human hand joint position. $\mathrm{E_j}$ is reported in centimeters.

\paragraph{Mean fingertip position error.}
The mean fingertip position error $\mathrm{E_{ft}}$ evaluates the fingertip-level imitation quality:
\[
    \mathrm{E_{ft}}
    =
    \frac{1}{T|\mathcal{F}|}
    \sum_{t=1}^{T}
    \sum_{f\in\mathcal{F}}
    \left\|
    x_{f,t}^{R}
    -
    x_{f,t}^{H}
    \right\|_2,
\]
where $\mathcal{F}$ denotes the fingertip set, and $x_{f,t}^{R}$ and $x_{f,t}^{H}$ are the positions of the $f$-th robot fingertip and reconstructed human fingertip at frame $t$, respectively. In our evaluation, $|\mathcal{F}|=5$. $\mathrm{E_{ft}}$ is reported in centimeters.

\paragraph{Success rate.}
The success rate $\mathrm{SR}$ measures the proportion of rollouts that satisfy the predefined thresholds on both object tracking and hand imitation errors. A rollout is considered successful if
\[
    \mathrm{E_r} < \tau_r,\qquad
    \mathrm{E_t} < \tau_t,\qquad
    \mathrm{E_j} < \tau_j,\qquad
    \mathrm{E_{ft}} < \tau_{ft}.
\]
By default, we set $\tau_r=30^\circ$, $\tau_t=3$ cm, $\tau_j=8$ cm, and $\tau_{ft}=6$ cm. The success rate is computed as
\[
    \mathrm{SR}
    =
    \frac{N_{\mathrm{success}}}{N_{\mathrm{rollout}}},
\]
where $N_{\mathrm{success}}$ is the number of successful rollouts and $N_{\mathrm{rollout}}$ is the total number of evaluated rollouts.
\clearpage

\end{document}